\documentclass{article}

\usepackage{arxiv}

\usepackage[utf8]{inputenc} 
\usepackage[T1]{fontenc}    
\usepackage{hyperref}       
\usepackage{url}            
\usepackage{booktabs}       
\usepackage{amsfonts}       
\usepackage{nicefrac}       
\usepackage{microtype}      
\usepackage{lipsum}
\usepackage{graphicx}
\usepackage{subfig}
\usepackage{algorithmicx}
\usepackage{amsmath}
\usepackage{amsfonts}
\usepackage{array} 
\usepackage[linesnumbered,ruled,vlined]{algorithm2e}

\hypersetup{
    colorlinks=true,
    linkcolor=blue,
    filecolor=blue,      
    urlcolor=blue,
    citecolor=blue
}

\usepackage[noend]{algpseudocode}
\usepackage{caption}

\algnewcommand\algorithmicforeach{\textbf{for each}}
\algdef{S}[FOR]{ForEach}[1]{\algorithmicforeach\ #1\ \algorithmicdo}

\title{Hyperdimensional Vector Tsetlin Machines with Applications to Sequence Learning and Generation}
\author{Christian D. Blakely
\\
  Centre for Artificial Intelligence Research \\
  University of Agder\\
  Grimstad, Norway \\
  \texttt{christian.blakely@uia.no}
}
\date{August 2024}

\hypersetup{
pdftitle={Hyperdimensional Vector Tsetlin Machines with Applications to Sequence Learning and Generation},
pdfsubject={Interpretable AI},
pdfauthor={Christian D.~Blakely },
pdfkeywords={Interpretable AI, Machine Learning, Dimension Reduction},
}

\begin{document}

\maketitle

\begin{abstract}
  We construct a two-layered model for learning and generating sequential data that is both computationally fast and competitive with vanilla Tsetlin machines, adding numerous advantages. Through the use of hyperdimensional vector computing (HVC) algebras and Tsetlin machine clause structures, we demonstrate that the combination of both inherits the generality of data encoding and decoding of HVC with the fast interpretable nature of Tsetlin machines to yield a powerful machine learning model. We apply the approach in two areas, namely in forecasting, generating new sequences, and classification. For the latter, we derive results for the entire UCR Time Series Archive and compare with the standard benchmarks to see how well the method competes in time series classification.
\end{abstract} \hspace{10pt}

\section{Introduction}

A large part of any design of a data learning agent is in feature extraction of the underlying data, and how it is computed and represented. The best processes for extracting features for learning information from data typically take advantage of expert knowledge of the underlying data to either expose the most relevant features, reduced noise, and extract the most amount of independent information in the data.  For many types of datasets, this might be challenging due to factors such as incoherence, abstractedness, or the sheer amount of noise present in the data. In designing features for Tsetlin machines, one is tasked to booleanize (or binarize) the underlying data, and under the presence of noise, this can be challenging. Furthermore, for notoriously complex high-dimensional data like noisy sequences, graphs, images, signal spectra, and natural language, creating encodings that are also interpretable for human reasoning in any post-hoc process can be difficult due to creating logic AND expressions that both take advantage of the relevant information in the data, but also lead to accurate expressions that can compete with other machine learning models.  

In this paper, we explore using Hyperdimensional Vector Computing (HV computing, or simply HVC) as an input layer to a novel Tsetlin machine architecture and apply it to learning, classifying, predicting, and generating sequences. Here, we argue that HVC can provide a robust layer of feature extraction due to the many computational advantages. This approach was first introduced in \cite{halenka2024exploringeffectshyperdimensionalvectors} and here, we streamline the approach to focus on sequences while further leveraging other attributes of HCV such as N-Gram sequence encoding and associative memory, while combining with TMs, to create a powerful hybrid methodology while remaining minimalist in memory sizes of the overall model. 

\subsection{Hyperdimensional Vector Representations}

There are many hyperdimensional vector (HV) representations that have been introduced for various applications over the past few decades. Combining with Tsetlin machines, the most natural approach is to use what is called in the literature as Binary Spatter Codes (BSC), or simply binary vectors of high dimension. These high-dimensional vectors, often referred to as HVs, are typically composed of thousands of bits. HVC is inspired by the way the human brain processes information, providing a robust and efficient means for computing and reasoning. Given the fact that Tsetlin machines learn directly from logical expressions on binarized data, BSC provides a natural input layer for Tsetlin machines.

While HVs are not typically treated as elements of a traditional vector space over a field, they share some properties with vector spaces. The algebra of binary HVs, which we will give an overview of in section \ref{hvc} is characterized by:

\begin{itemize}
\item \textbf{Bundling:} Combining multiple vectors into one using majority voting or similar methods.
\item \textbf{Binding:} Associating vectors using the XOR operation.
\item \textbf{Unbinding:} Retrieving original vectors from bound vectors using the XOR operation.
\item \textbf{Similarity Measurement:} Using Hamming distance to measure the similarity between vectors.
\item \textbf{Perturbation:} Slightly modifying vectors to represent related but distinct concepts.
\end{itemize}
These operations form the core algebraic framework for HVC, enabling robust and efficient computation in binary spaces

\subsection{Motivation and contributions}

One of the biggest attractions of HVC is the ability to represent virtually any type of data through the use of the vector operations. The main motivation in this paper is to build a strategy for learning and generating sequences by establishing a robust encoding from multidimensional sequences to HVs.  We also desire that the encoding is then naturally suited for learning with TMs, and so we will provide empirical evidence of an attractive coupling of both HVC and TMs.  

This paper contributes a new approach to sequence learning, where we show we can compete with current SOTA results on fundamental benchmarks. Furthermore, we also show how combining some of the features of HVC with TMs can produce a highly versatile forecasting model for many types of sequences, while generating new sequences of any length with are similar in Hamming distance to any given sequence. Lastly, we will show that the models have a very light footprint in memory, making it an attractive approach for online learning in small embedded systems.

\subsection{Organization of Paper}

We first give a brief overview of HVC along with the core algebraic framework we will be using throughout the paper. We then discuss the encoding strategy we will be using for various types of sequences, along with designing a so-called associative memory which will be very useful in sequence generation and forecasting.

The next section briefly outlines the TM architecture we will being using throughout, with references to more in-depth treatments of the approach. 

We will present an in-depth numerical treatment of the approach by first outlining the hybrid structure of our HVC-TM architecture, and then providing an algorithm for time series learning and forecasting. In the final section of the paper, we will provide results for applying our proposed approach on the entire UCR Time Series archive for classifying many types of time series. 

Finally, we will conclude the study with some alternatives in design and further next steps with possible applications. 

\section{Review of Binary HVs}\label{hvc}

We will leverage heavily throughout our approach some fundamentals of HVs. A fantastic survey and introduction to HVC can be found in the recent monograph \cite{Kleyko_2022}, which also gives insights into current research trends. To give a quick overview on why one would want to use HVs   
\begin{itemize}
  \item \textbf{Robustness}: HVs are resilient to noise and errors. Small perturbations in the vectors do not significantly affect their overall similarity, making HVC robust to noisy data.
  \item \textbf{Scalability}: HVC can easily handle high-dimensional data, and the operations on HVs are computationally efficient.
  \item \textbf{Simplicity}: The operations on HVs, such as bundling, binding, and similarity measurement, are straightforward and can be implemented efficiently.
  \item \textbf{Flexibility}: HVs can represent a wide range of data types and structures, from simple scalars to complex symbolic representations.
\end{itemize}

The operations we use of binary HVs differ from standard real vectors in that we don't use multiplication and addition. Instead, we apply operations defined by binding and bundling on HVs, followed by the inclusion and notion of a distance for comparing HVs. Perturbations of vectors are also used to represent the notion of ordering in HVs (for example, vector $\mathbf{A}$ comes before $\mathbf{B}$ in a sequence). Here we give a brief introduction to these operations

\subsection{Bundling}

Bundling is the operation of combining multiple HVs into a single HV. This operation is analogous to the concept of superposition in physics, where multiple states are combined. In HVC, bundling is typically implemented using element-wise majority voting and will be represented throughout the paper as a $+$.

Given binary HVs $\mathbf{A}$, $\mathbf{B}$, and $\mathbf{C}$, the bundled HV $\mathbf{S}$ can be expressed as $\mathbf{S} = \text{Majority}(\mathbf{A}, \mathbf{B}, \mathbf{C})$ where for each bit position $i$:
\begin{equation}
S_i = \begin{cases} 
1 & \text{if } A_i + B_i + C_i \geq 2 \\
0 & \text{otherwise.}
\end{cases}
\end{equation}
Because of the fact we take a majority, it is important that we only apply this operation on an odd number of HVs. 

\subsection{Binding}

Binding is the operation of combining two HVs to form a new HV that represents their association. In HVC, binding is typically implemented using the element-wise XOR operation. Given binary HVs $\mathbf{A}$ and $\mathbf{B}$, the bound HV $\mathbf{C}$ is:
$\mathbf{C} = \mathbf{A} \oplus \mathbf{B}$
where $\oplus$ denotes the XOR operation for each bit position $i$, $C_i = A_i \oplus B_i$
The unbinding operation is used to retrieve an original HVs that was bound together with another, and is frequently used in decoding HVC representations. This is achieved by applying the XOR operation with one of the original HVs to the bound HV. 

Given the bound HV $\mathbf{C}$ and one of the original HVs (say, $\mathbf{A}$), we can retrieve the other HV ($\mathbf{B}$) as follows:
$\mathbf{B} = \mathbf{C} \oplus \mathbf{A}$
This works because the XOR operation is its own inverse. Specifically, $\mathbf{A} \oplus \mathbf{A} = \mathbf{0} \quad \text{(the zero vector)}$
giving
$\mathbf{C} \oplus \mathbf{A} = (\mathbf{A} \oplus \mathbf{B}) \oplus \mathbf{A} = \mathbf{A} \oplus \mathbf{A} \oplus \mathbf{B} = \mathbf{0} \oplus \mathbf{B} = \mathbf{B}$

Thus, by XORing the bound HV $\mathbf{C}$ with one of the original HVs ($\mathbf{A}$), we effectively cancel out $\mathbf{A}$ and retrieve the other original HV ($\mathbf{B}$). For example, suppose we have two binary HVs $\mathbf{A}$ and $\mathbf{B}$:
$
\mathbf{A} = [1, 0, 1, 1], \quad \mathbf{B} = [0, 1, 0, 1]
$
Their bound HV $\mathbf{C}$ is $
\mathbf{C} = \mathbf{A} \oplus \mathbf{B} = [1 \oplus 0, 0 \oplus 1, 1 \oplus 0, 1 \oplus 1] = [1, 1, 1, 0]
$
To retrieve $\mathbf{B}$, we unbind $\mathbf{C}$ with $\mathbf{A}$:
$
\mathbf{B} = \mathbf{C} \oplus \mathbf{A} = [1 \oplus 1, 1 \oplus 0, 1 \oplus 1, 0 \oplus 1] = [0, 1, 0, 1]
$
We will often compare HVs, and this is done via a similarity measure called the Hamming distance which is applied between two binary vectors $\mathbf{A}$ and $\mathbf{B}$ of equal length is defined as the number of positions at which the corresponding bits differ, namely $d_H(\mathbf{A}, \mathbf{B}) = \sum_{i=1}^{n} \left( A_i \oplus B_i \right)$. The Hamming distance is a measure of dissimilarity between two binary vectors, so a Hamming distance of 0 indicates that the vectors are identical, while a larger Hamming distance indicates greater dissimilarity. 

The final operation useful for sequences is perturbation operator. It is the operation of slightly modifying a HV to represent a related but distinct vector. We implement this by simply shifting cyclically the bits of the vector $j$ places. We will denote a perturbation applied to a vector $hv$ by $p(hv, j)$. The inverse operation of perturbation $ip$ is then just $p(hv, -j)$. This will be used for representing the $j$th position in a sequence. We will typically call the vector resulting in applying a $j$ perturbation on a vector as a position vector. 

\subsection{Sequence operations}

We can now apply these operations for representing and processing sequences of data. Several recent studies have been proposed for encoding sequences, such as in \cite{schlegel2022hdcminirocketexplicittimeencoding} and \cite{rachkovskij2022recursivebindingsimilaritypreservinghypervector}. In a similar fashion, the goal of our approach will be to encode a time series sequence $\mathbf{s} \in \mathrm{R}^n$ into a binary vector dimension of $D$. In this paper, we will represent sequences (or time series) as a stationary sequence of scalars $\mathbf{s}  = [s_1, s_2, \ldots, s_n]$. We will show how we represent these as HVs, leveraging the operations of bundling, binding, and perturbation to capture the structure and relationships within the sequence.

The first step is to introduce what we call an Interval embedding. Given a quantization of dimension $Q$, since we assume our sequences are stationary, they are bounded and thus have a minimum and maximum interval, $[m_0, m_1]$. We divide the interval $[m_0, m_1]$ into $Q$ buckets, can represent each bucket by an HV (we will denote these as $hv$. Thus the process of mapping $S = [s_1, s_2, \ldots, s_n]$ into a collection of HVs $E_S = [hv_1, hv_2, \ldots, hv_n]$ will be represented as $E$. This is shown in the following algorithm.

\begin{algorithm}[H]
\SetAlgoLined
\KwIn{Scalar value $x$, Interval range $[m_0, m_1]$, Number of divisions $Q$}
\KwOut{Hyperdimensional vector $hv$}
\BlankLine
Initialize $step \leftarrow \frac{m_1 - m_0}{Q}$\;
Initialize $intervals \leftarrow \{\}$\;
\For{$i \leftarrow 0$ \KwTo $Q-1$}{
    $intervals[i] \leftarrow \text{Random hyperdimensional vector}$\;
}
$index \leftarrow \left\lfloor \frac{x - m_0}{step} \right\rfloor$\;
$hv \leftarrow intervals[index]$\;
\Return $hv$\;
\caption{Interval Embedding}
\end{algorithm}

And now we use the interval embedding function to encode the original sequence into a sequence of HV interval embeddings.

\begin{algorithm}[H]\label{encodingScalar}
\SetAlgoLined
\KwIn{Scalar sequence $S = [s_1, s_2, \ldots, s_n]$, Interval embedding function $E$}
\KwOut{Encoded sequence $E_S = [hv_1, hv_2, \ldots, hv_n]$}
\BlankLine
$E_S \leftarrow \{\}$\;
\For{$i \leftarrow 1$ \KwTo $n$}{
    $hv_i \leftarrow E(s_i)$\;
    $E_S \leftarrow E_S \cup \{hv_i\}$\;
}
\Return $E_S$\;
\caption{Encoding Scalar Sequence}
\end{algorithm}

With any sequence now being transformed into a sequence of HVs, the next step is encoding this sequence into one HV. Here we apply both perturbations and binding with N-Grams. 
To encode the original sequence $E_S = \{hv_1, hv_2, hv_3, \ldots, hv_n\}$, apply a convolution comprised of the perturbation representing a position vector applied to the $hv_i$ followed by a binding with a Gram vector $p(\cdot, j) \oplus G_j$. If the number of Gram vectors is, for example 3, the first element would be
$GV_2 = (p(hv_0, 0) \oplus G_0) \oplus (p(hv_1, 1) \oplus G_1) \oplus (p(hv_2, 2) \oplus G_2)$. This can be summarized below.
\paragraph{Binding with Position Vectors}
\begin{itemize}
  \item Generate random position vectors for each position in the N-Gram $G_j$, $j=1,\ldots,N$. 
  \item Bind each element of the N-Gram with its corresponding position vector using the XOR operation.
\end{itemize}
For short, we will define the N-Gram vector at the $t$-th temporal index by
$GV_t = (p(hv_t, 0) \oplus G_0) \oplus (p(hv_{t-1}, 1) \oplus G_1) \oplus (p(hv_{t-N}, N) \oplus G_N)$
With the position encoded, and the N-gram contains the spatial information
of N subsequent samples with different timestamps, making it a spatiotemporal HV.

\begin{itemize}
  \item Divide the sequence into overlapping N-Grams. For example, for a sequence $E_S = \{hv_1, hv_2, hv_3, \ldots, hv_n\}$ and $N = 3$, the N-Grams are $\{hv_1, hv_2, hv_3\}$, $\{hv_2, hv_3, hv_4\}$, etc.
  \item Apply the permutation operator $p(\cdot,j)$ to the $j$th position, yielding a position vector $P_j := p(hv, j)$
  \item At each position, bind the position vector with the $j$th Gram HV
  \item Bind all the N-Gram position vectors together
\end{itemize}
Once we have generated $GV_0, \ldots, G_{L - N -1}$, where $L$ is the length of the sequence and $N$ is the number of grams, then we can then apply a bundling operation to combined the elements in the HV space. Recall that bundling the bound HVs of the N-Grams using element-wise majority voting. In summery, the steps of encoding a sequence is shown if the figure below.

\begin{figure}[!ht]
\centering
\includegraphics[width=.7\textwidth]{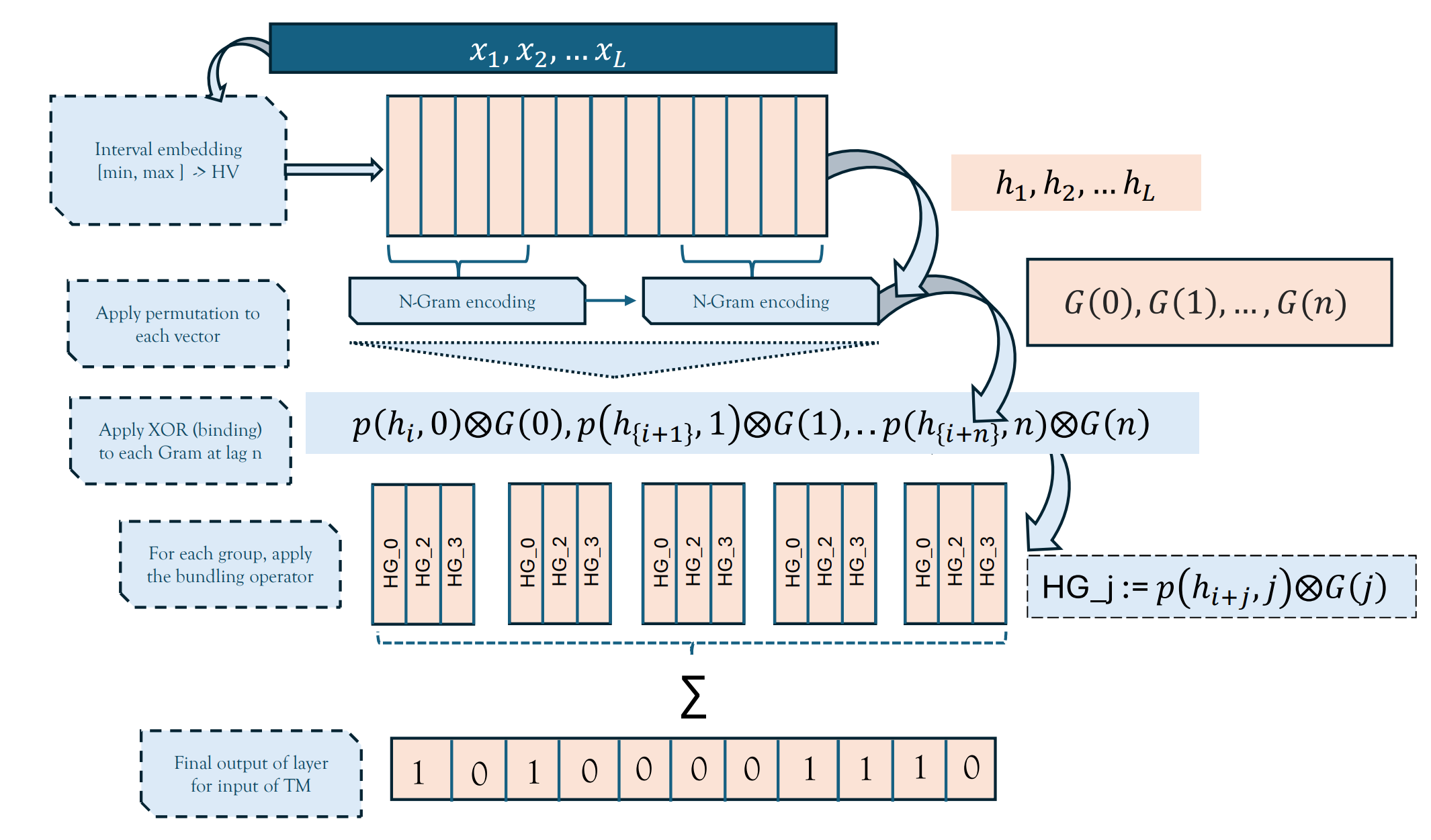}
\caption{The procedure for encoding a sequence into a hypervector. }
\label{ref:encoding}
\end{figure}

The first step in the top encodes all the scalar values of the entire sequence of time series values into a sequence of HVs based on the interval embedding procedure. In order to encode these HVs into a single HV, we apply N-Gram encoding by first applying a permutation to the respective position giving encoding position, and then we proceed by apply XOR operation with the N-Gram vectors. This is essentially a convolution across the time series sequence given not with temporal and spatial information. With the collection $L - N - 1$ HVs post XORing with N-Gram vectors, we then apply a bundle operation to store all the information in one HV. This vector is now ready as input into a TM model, along with a class label or the next value in the sequence (forecasting). 

We now apply this procedure to build an associative memory for a collection of sequences that will be used for our hybrid forecasting approach.

\subsection{Associative Memory}

Associative memory in the context of HVC refers to a memory model where stored items are retrieved based on their similarity to a query item. It leverages the properties of high-dimensional spaces to perform efficient and robust retrieval operations. We propose a hybrid model of forecasting that is a multistep step approach based on associative memory. 

To build the associative memory for a given set of sequences, the process involves taking all possible N-Grams of the series and encoding them into HVs, while storing them with their associative values in the sequence. More formally, for a set of $M$ sequences $\mathbf{S} = \{ S_0, \ldots, S_M \}$, where $S_i := s_0, s_1, \ldots, s_L$, we extract all possible N-Grams from each sequence. We will define 
\begin{equation}
\mathbf{H}^m_{t} := (p(hv^m_t, 0) \oplus G_0) \oplus (p(hv^m_{t-1}, 1) \oplus G_1) \oplus (p(hv^m_{t-N}, N) \oplus G_N) 
\end{equation}
by the $t$th N-Gram HV from the $m$th sequence, and then define the mapping
$$
\mathcal{M} (\mathbf{H}^m_{t}) \mapsto s^m_t,
$$
for all sequence values $s^m_t$. This creates a dictionary, namely a (key, value) pairing where $\mathcal{M}$ is the Memory, the key is a HV, and the value is the interval embedding HV representing a scalar. The process of querying this mapping is called retrieving memory from the associated memory. To retrieve an item similar to a query HV $\mathbf{Q}$, the memory HV $\mathbf{M}$ is queried, and the stored HV $\mathbf{H}$ that is most similar to $\mathbf{Q}$ is retrieved. The similarity is measured using the Hamming distance:
\begin{equation}\label{eq:hamming}
\mathbf{H}_{\text{retrieved}} = \arg\min_{\mathbf{H} \in \text{Memory}} d_H(\mathbf{Q}, \mathbf{H})
\end{equation}
Given a set of stored HVs $\{\mathbf{H}_1, \mathbf{H}_2, \ldots, \mathbf{H}_k\}$, the memory HV $\mathbf{M}$ is:
$
\mathbf{M} = \sum_{i=1}^{k} \mathbf{H}_i
$
To retrieve the item most similar to the query $\mathbf{Q}$, we calculate the Hamming distance between $\mathbf{Q}$ and any proposed $\mathbf{H}_i$. This process allows for efficient and robust retrieval of stored items based on their similarity to the query item. Before we describe the forecasting and classification approach of HVTMs, we first review the TM architecture we will be using.

\section{Tsetlin Machine Architecture}

\begin{figure}[!ht]
\centering
\includegraphics[width=.7\textwidth]{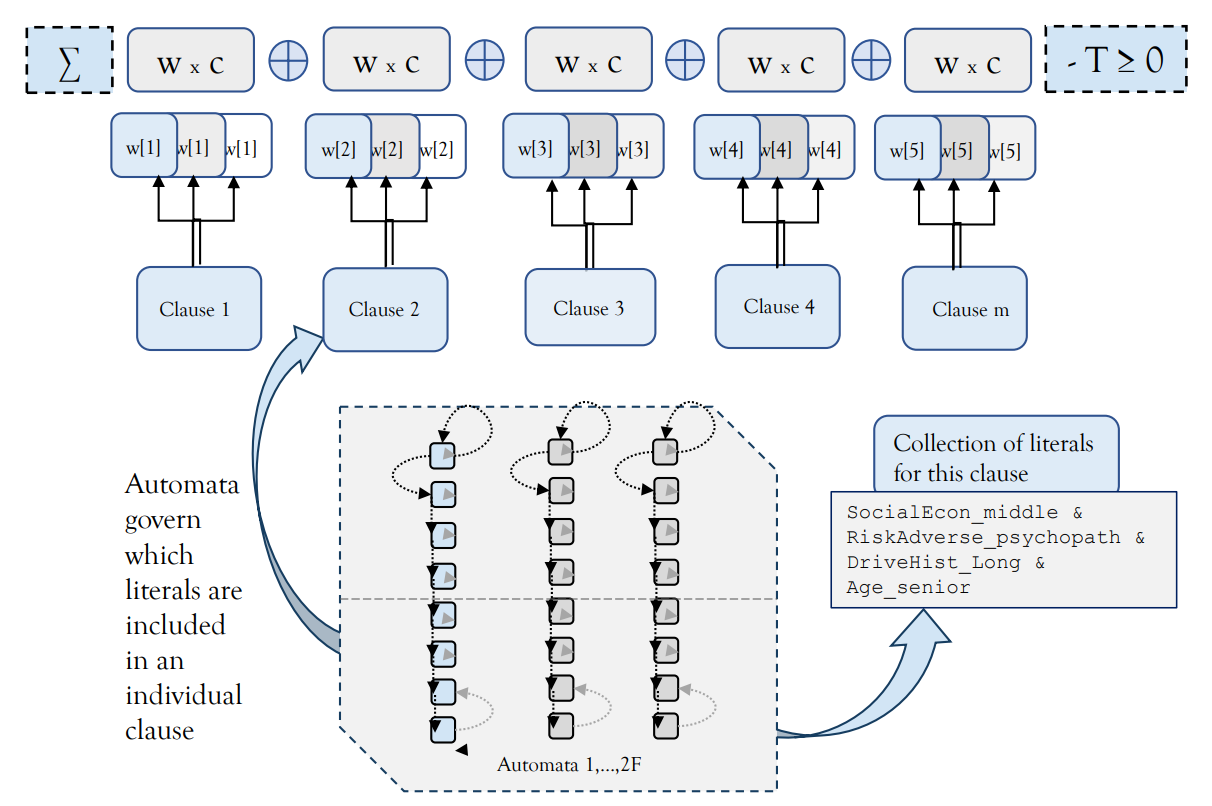}
\caption{Showing a collection of clauses and their relationship with weights that are learned during TM training. Each clause contains a set of literals that are also learned during feedback.}
\label{ref:clause-weights}
\end{figure}

In this paper we employ a recently developed architecture for TM learning originally proposed in \cite{glimsdal2021coalescedmultioutputtsetlinmachines}. The architecture shares an entire pool of clauses between all output classes, while introducing a system of weights for each class type. The learning of weights is based on increasing the weight of clauses that receive a Type Ia feedback (due to true positive output) and decreasing the weight of clauses that receive a Type II feedback (due to false positive output). This architectural design allows to determine which clauses are inaccurate and thus must team up to obtain high accuracy as a team (low weight clauses), and which clauses are sufficiently accurate to operate more independently (high weight clauses). The weight updating procedure is given in more detail in \cite{glimsdal2021coalescedmultioutputtsetlinmachines}. Here we illustrate the overall scheme in Figure \ref{ref:clause-weights}. Notice that each clause in the shared pool is related to each output by using a weight dependant on the output class and the clause. The weights that are learned during the TM learning steps and are multiplied by the output of the clause for a given input. Thus clause outputs are related to a set of weights for each output class. 

An additional parameter we use for our TM architecture is to ensure that a maximum number of literals is allowed for every clause. This has the effect of ensuring that the clauses do not become too "dense", proposed in \cite{ostby2024sparsetsetlinmachinesparse}.

\begin{figure}[!ht]
\centering
\includegraphics[width=.7\textwidth]{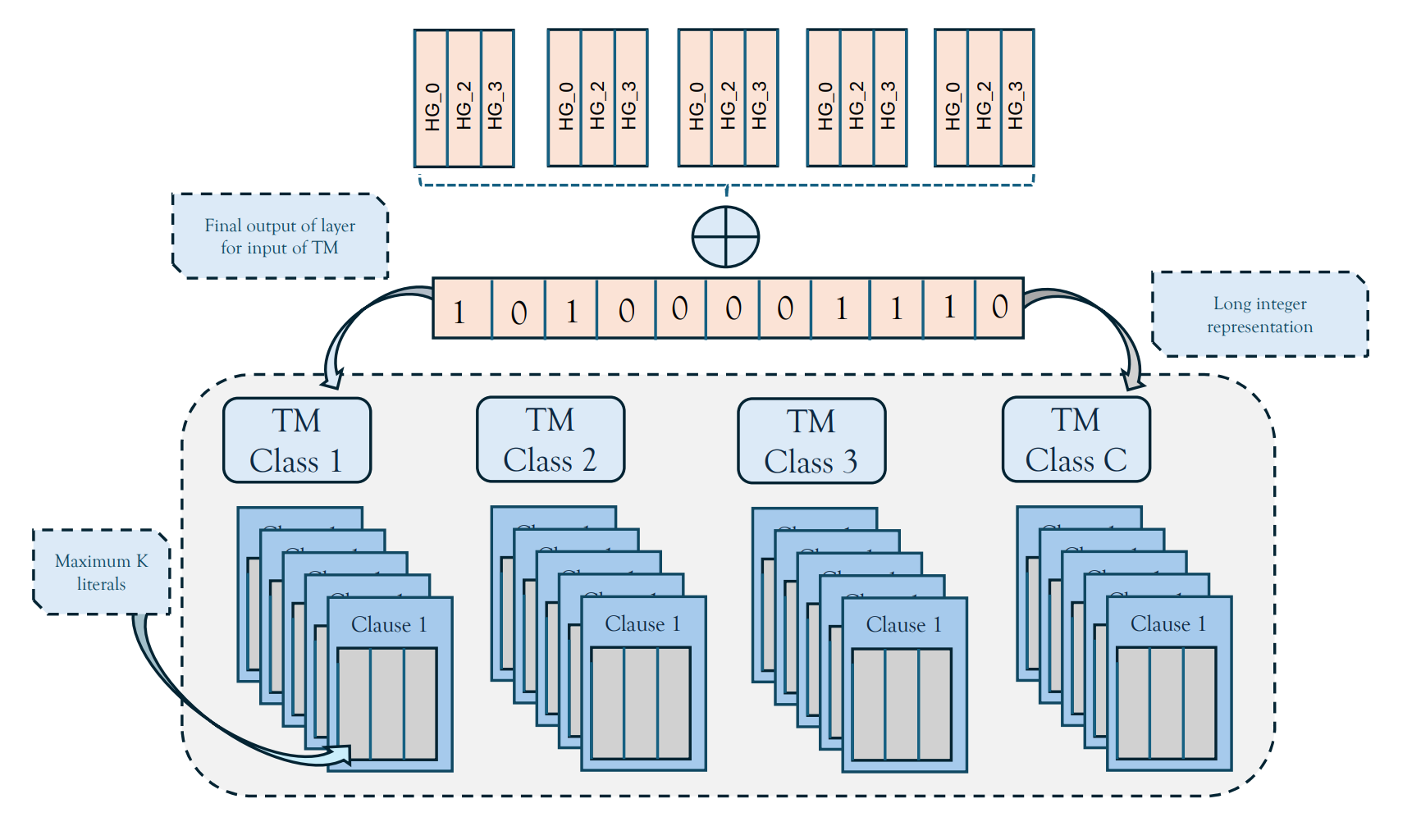}
\caption{Input into the TM is the N-Gram encoded latest $N$ values of the sequence. This input is used to predict the next value governed by the $Q$-class TM}
\label{ref:tsetlinMachine}
\end{figure}

\section{Time Series Classification}

In this section we do an in-depth study on how the HVTM performs in classifying many types of time series. For this study, we apply our proposed methodology on the UCR Time Series Classification archive \cite{UCRArchive2018} which is a widely used comprehensive collection of datasets for bench-marking time series classification algorithms. Boasting 128 different time series datasets from a broad range of domains, including medicine, biology, finance, motion tracking, and sensor data with a diverse range of time series lengths, some as short as a few dozen with very few training samples or having a high imbalance of a certain class, all while having strong benchmark tracking, it is the premier data archive for testing new time series classification models. 

Here we apply our methodology to all 128 data sets, and for this we fix the HV strategy, namely setting
\begin{itemize}
\item HV dimension = 5000
\item N-Gram length = 3
\item Quantization = 50
\end{itemize}

For the TM architecture, we randomly choose 20 different configurations where
\begin{itemize}
\item number clauses between [100, 2000]
\item number max literals per clause [10, 100]
\item specificity between [10f, 20f]
\item threshold between [0, 100].
\end{itemize}
For each random configuration, we run on 10 Epochs, and take the accuracy after the final Epoch. Finally, to measure the final performance, we then take the top 10 model configurations and report the maximum accuracy.

\subsection{Numerical Results}

In order to compare our approach with the published benchmarks of \cite{UCRArchive2018}, for all 128 time series, we compare the accuracy of our classification with the best accuracy recorded in \cite{UCRArchive2018}. Their approach compares Euclidean distance, and two Dynamic Time Warping (DTW) computations (one with a learned/optimized parameter warping parameter, and one with a fixed parameter).  

To visually compare, we plot the pair ($\text{accuracy}_{HVTM}, \text{accuracy}_{UCR}$).

Without having chance for optimization on controlling the dimension of the HV, nor the number of Grams used to encode the hypervector, and using a random parameterization for the TM, we can see that the approach has performed quite well simply using the "out-of-the-box" default settings.  The HVTM method improves or competes in accuracy (cutoff at 2 percent) of the optimal benchmark provide by \cite{UCRArchive2018} in roughly 78 percent of the data sets.

\begin{figure}[!ht]
\centering
\includegraphics[width=.8\textwidth]{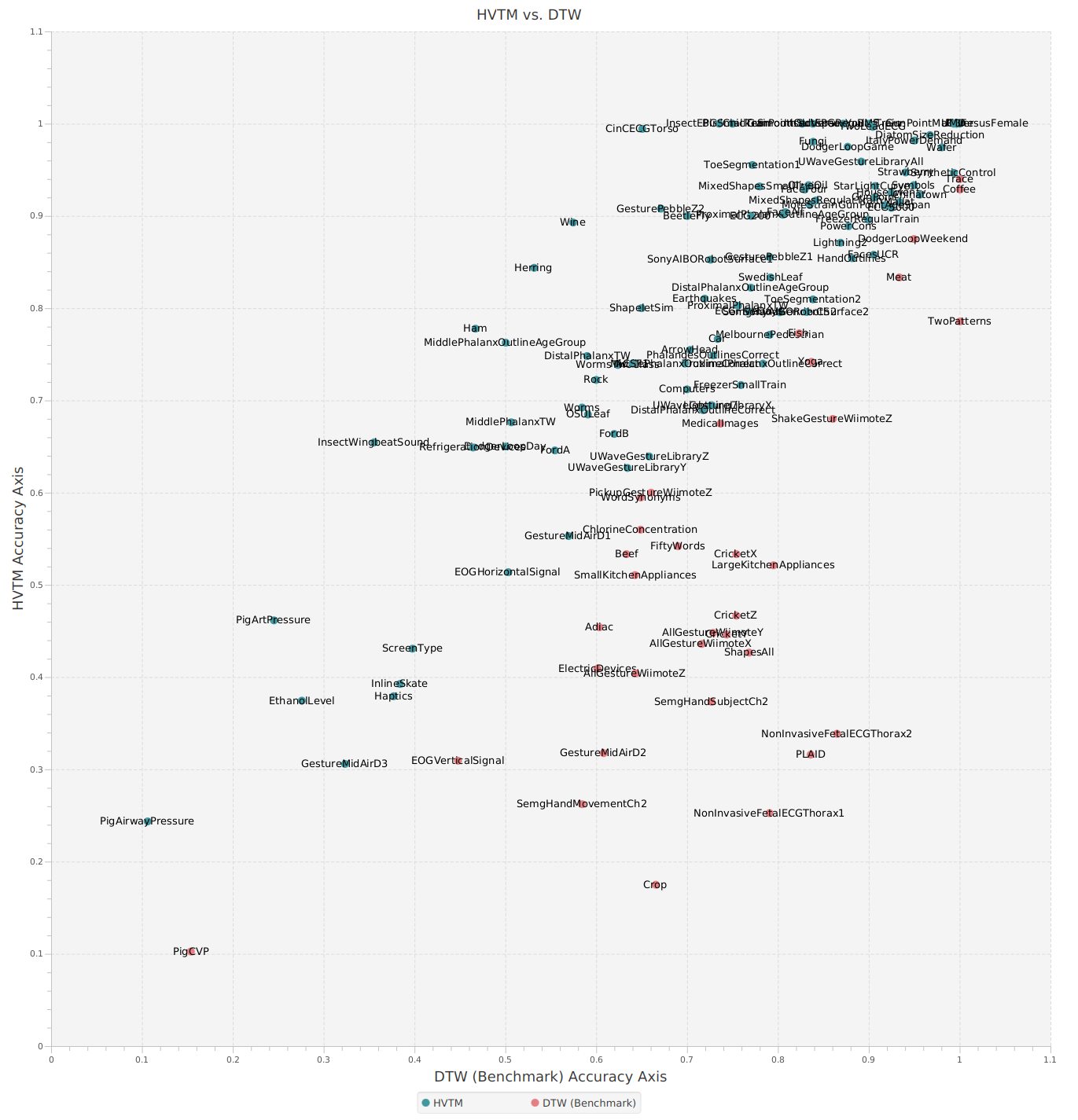}
\caption{Scatter plot showing the accuracy of the performance on each data set versus the benchmark. Here the X-axis is the benchmark accuracy and the Y-axis is the HVTM accuracy}
\label{ref:scatterPlot}
\end{figure}

To give an overview of the performance comparison, we employ a scatter plot showing the accuracy of the HVTM approach on each data set versus the DTW benchmark. In Figure \ref{ref:scatterPlot}, the scatter plot shows the clustering along $x=y$ line, demonstrating that the approach is very competitive with the benchmark. 

\begin{figure}[!ht]
\centering
\includegraphics[width=.8\textwidth]{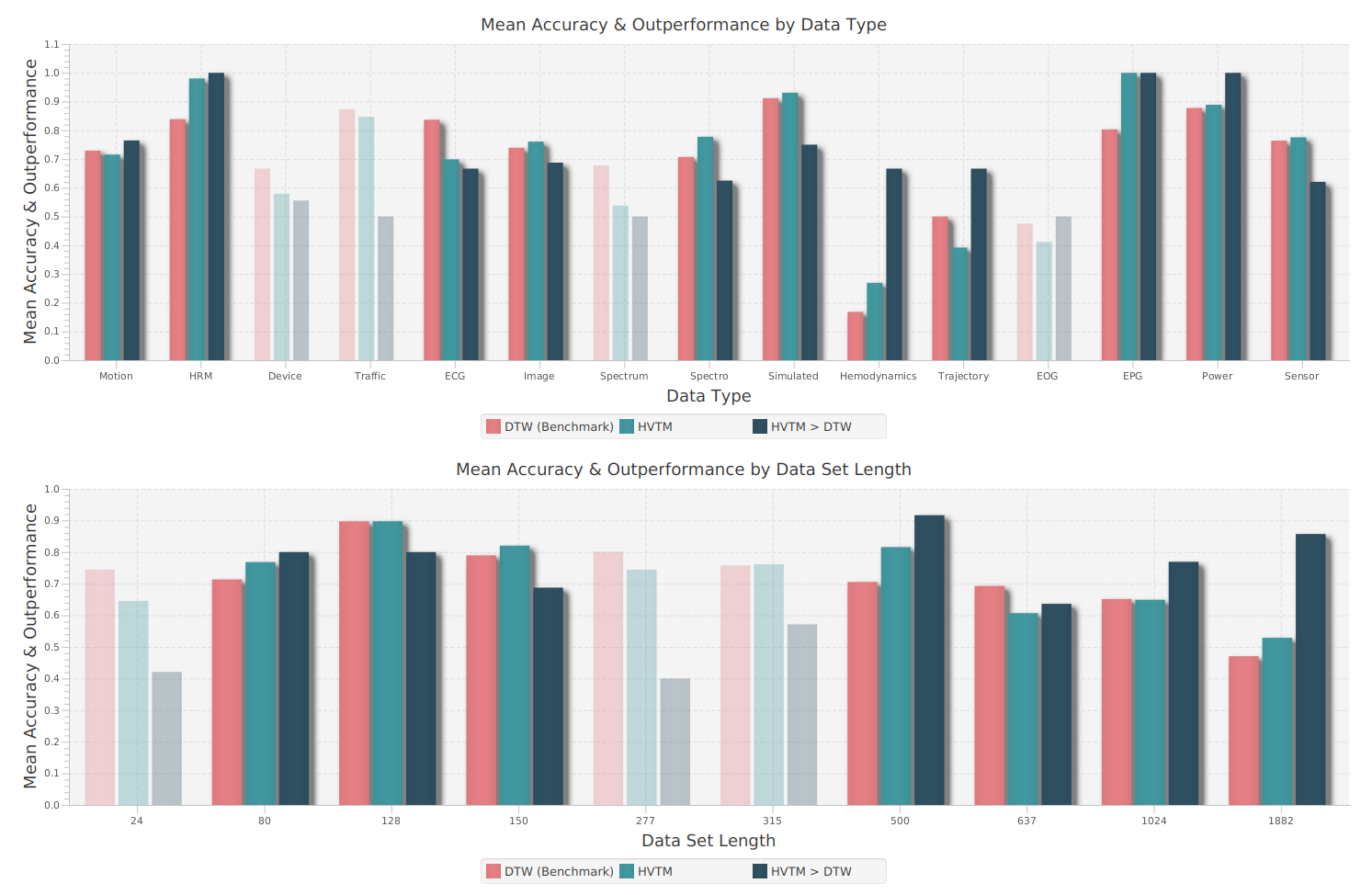}
\caption{Mean accuracy and (out)performance results into four categories: data type, length of time series}
\label{ref:dataTypeComparison}
\end{figure}

To get even more insights into the (out)performance, we break down further the results into four categories: data type, length of time series, number of training samples available, and number of classes in order to gain possible information on where HVTMs outperformance different training setups. Figure \ref{ref:dataTypeComparison} shows the comparison of performance across the different data types and the length of the time series.  The first two bars in each category represent the mean accuracy for both HVTM and the DTW benchmark. The third bar show the percentage of times HVTM outperformed the benchmark in terms of accuracy. We highlight (using a shadow on the bars) the categories in which HVTM outperformed DTW at least 60 percent of the time. 

We can see that HVTM outperformed the DTW benchmark on most of the data set types, including Motion, Images, and ECGs. A few categories, including EOG, and Traffic, only two data sets fall into these category making performance comparisons difficult. 

Next we see that if we compare the performances broken down into time series length, there is clear dominance by HVTM over most lengths. However, it seemed to struggle with very short series, namely series of length 24-80. Furthermore, for mid-range series, 277-500 in length, there is no clear outperformance. However, for series greater than 500 in length there seems to be a clear dominance by the HVTM approach.

We also are interested in how the methods compare given the number of training samples. More training samples can yield more dispersion of information for each class, but here HVTM has no problem in outperforming DTW benchmarks for a very few number of training samples. However, DTW seems to outperform for a very large numbe or training samples (more than 1000). In regards to number of classes, HVTM can handle less than 10 classes very well, and outperforms the DTW approach fairly well. However, for many more classes, the HVTM approach with the given HV parameters seems to struggle. This could be simply due to only utilizing a dimension of 5000 for the HV system. 

\begin{figure}[!ht]
\centering
\includegraphics[width=.8\textwidth]{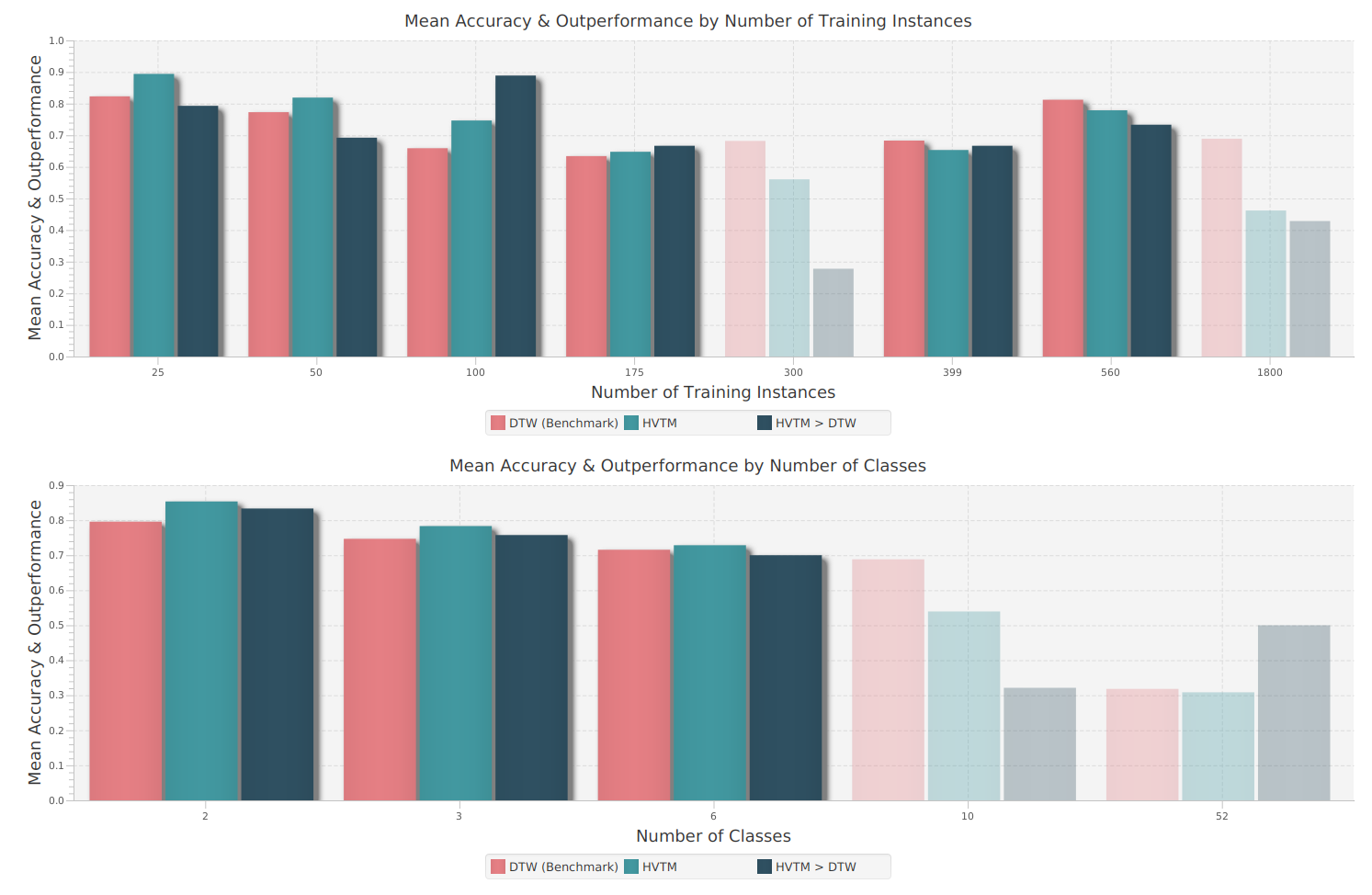}
\caption{Mean accuracy and (out)performance results into four categories: number training samples, number classes}
\label{ref:trainingComparison}
\end{figure}

In most categories across all 4 bar charts of performance comparisons, there does not exist a large discrepancy between average error rates. The only instances of any large discrepancy is for a large number of classes. To improve on this, some further optimization on either the dimension $D$ of the HV space, the number of N-Grams, or allowing more candidate TM models might improve the accuracy of the HVTM approach. One could also consider encoding more information into the HV, such as the class label (or potential class label) and first training purely in the HV space by using associative memory. 

\subsection{Hybrid Predictions}

With the computational framework for both HVC and our TM architecture, we now show a straightforward approach to forecasting and generating new sequences based on encoded historical sequences. Prediction involves using the encoded HV of a sequence to predict the next element(s) in the sequence. Here we make a few assumptions on the underlying time series data
\begin{itemize}
    \item The time series data is stationary, namely we can assume there is a lower and upper bound indefinitely during any time range we consider
    \item While only one time series is necessary and sufficient, any additional time series for training purposes can be considered from the the same data generating process
    \item No NAs/missing values exist in the time series, namely all the data for training is available
\end{itemize}
We also note that we make no assumption on the uniformity of the data sampling of the underlying time series. So all series are considered to have uniform time stamps. However, some applications could utilize the distance between each time stamp as a feature, but we will not consider this in the study.

Our forecasting approach considers two steps for training: 1) encoding the time series into a dictionary of N-Grams mapped to time series values by creating an associative memory, and 2) Training a TM on the dictionary of N-Gram HVs labeled with the final sequence value in that N-Gram.  

\begin{itemize}
    \item Given the latest observation $s_n$ in the time series, find the next observation $\hat{s}_{n+1}$ that has the highest similarity from the associative memory based on the current context $s_{n-N}, \ldots, s_n$ 
    \item Use associative memory to retrieve the most similar past context and predict the next element based on this context.
    \item For the proposed observation $\hat{s}_{n+1}$, recursively generate then next $\hat{s}_{n+2}, \hat{s}_{n+3}, \ldots, \hat{s}_{n+K}$
    \item For each proposed observation $\hat{s}_{n+1}$, using the TM with $Q$ classes, where $Q$ is the quantization level, predict the each next observation $\bar{s}_{n+1+i}$ given the gram vector $G$
    \item Combine the forecasts $\hat{s}_{n+1} + \bar{s}_{n+1}$
\end{itemize}

\begin{figure}[!ht]
\centering
\includegraphics[width=.7\textwidth]{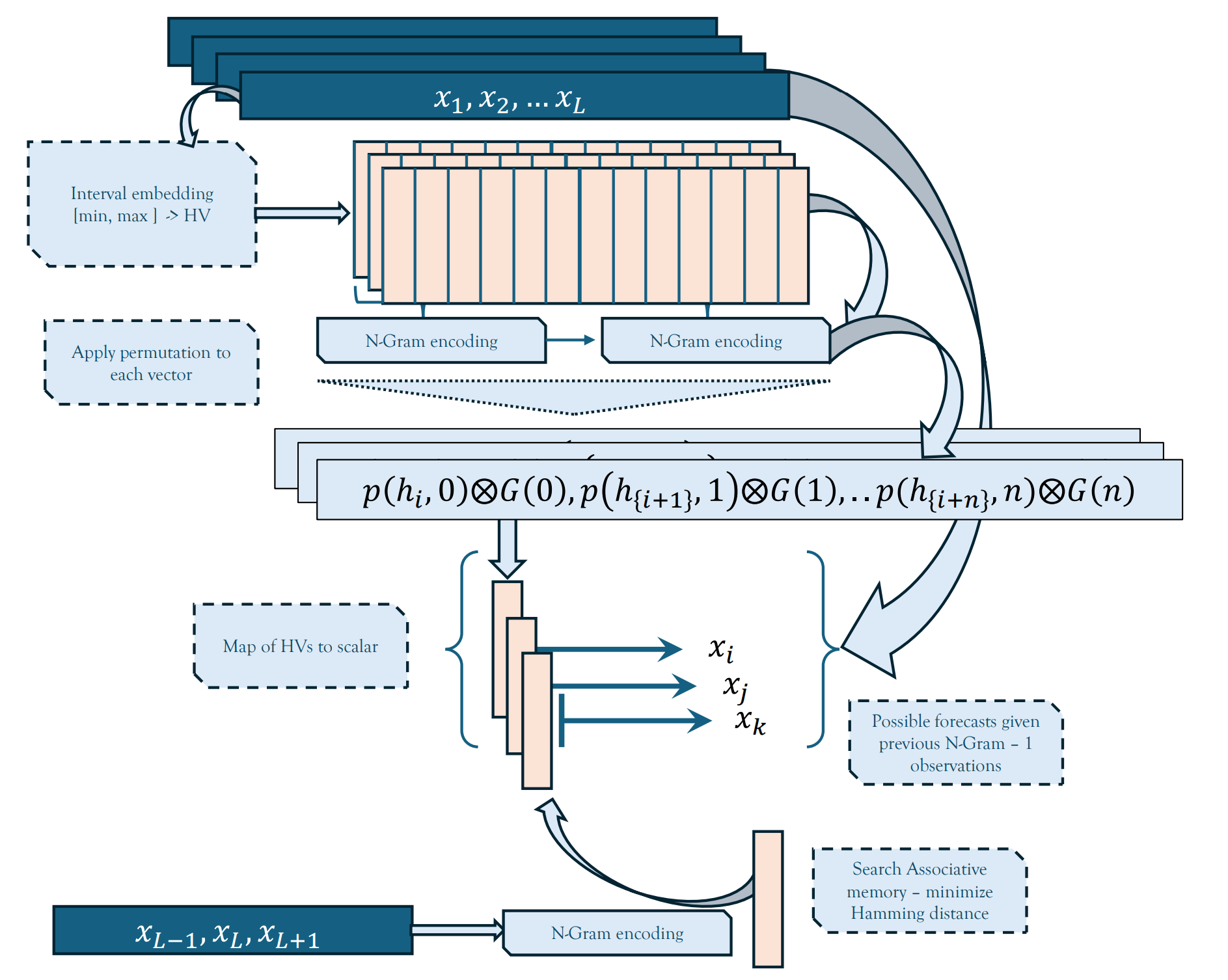}
\caption{Showing the steps for encoding based on N-Gram encoding, then forecasting given the current encoded vector based on the two most recent observations.}
\label{ref:forecasting}
\end{figure}

The second step is done by using as input into the HVTM the N-Gram encoded vector of the last $N$ values of the time series, followed by predicting the next quantized value in the sequence. This gives us a second forecast, from which we can derived a weighted average from the associative memory forecast. 

\begin{figure}[!ht]
\centering
\includegraphics[width=.7\textwidth]{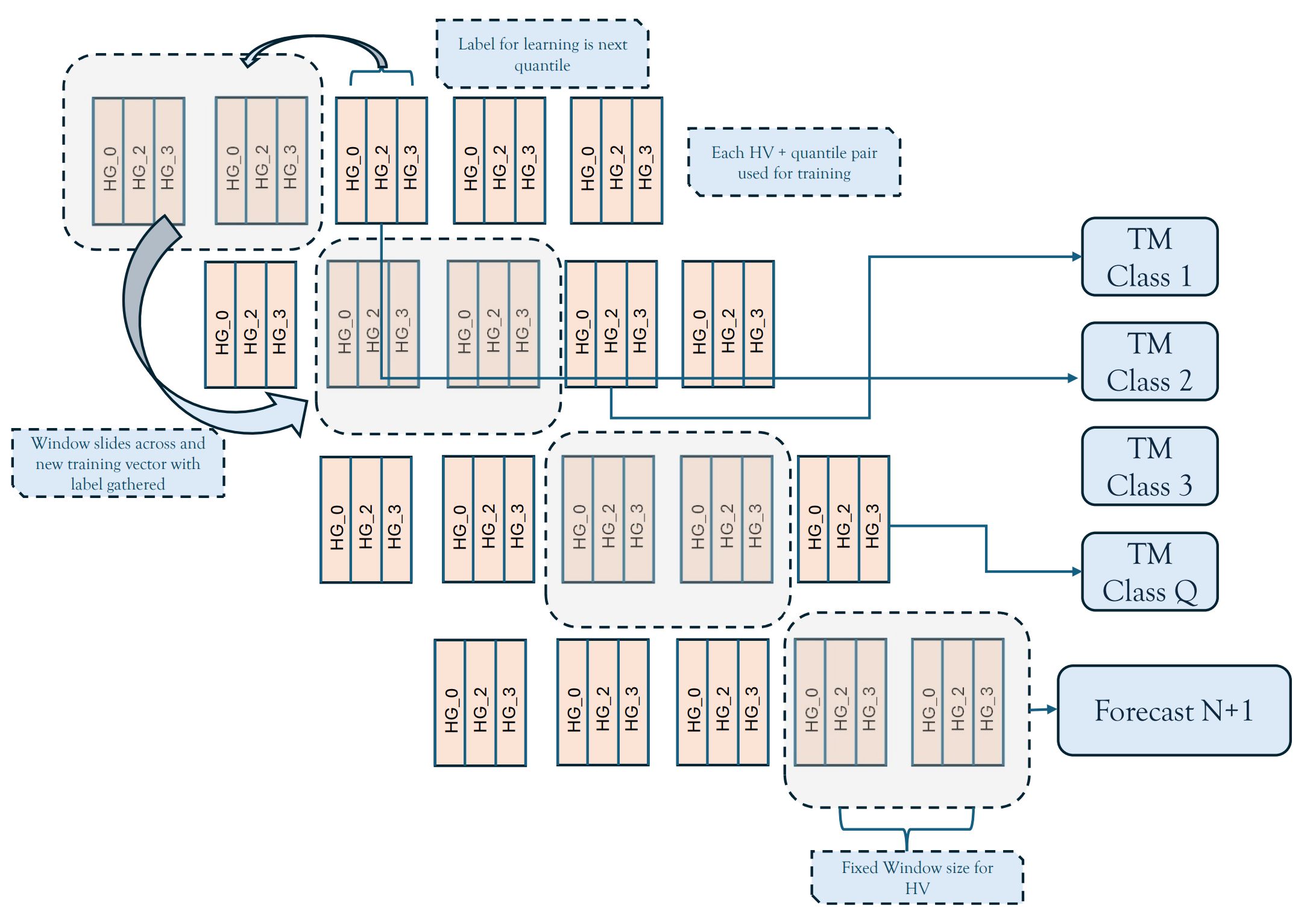}
\caption{Input into the TM is the N-Gram encoded latest $N$ values of the sequence. This input is used to predict the next value governed by the $Q$-class TM}
\label{ref:hybridForecast}
\end{figure}

This is shown in Figure \ref{ref:hybridForecast}. Each block of window size $M$ leads to a group of HVs that is the collection of N-Gram encoding of length $N$. This HV is used as the data with the label being the quantized value of the next value in the sequence. This procedure produces many labeled samples from which we can train the TM using the HVs. For prediction, we combine the prediction made by using the latest HV in the time series sequence and combine it with the prediction made by the TM model.

\subsubsection{Numerical Experiments}

In order to empirically determine how well the forecasting and sequence generation performs, we begin by simulating deterministic sequences and comparing different length forecasts with the simulated sequence. For this, our model is a harmonic series 
$s(t) = a \sin( t 2 \pi) * \cos( b t 2 \pi + ct) + \sin(2 \pi + t)$ where we randomly choose $a, b, c \in [0, 1]$ to generate a sequence. For our model setup, we again use $D=5000$ and the number of N-Grams we set at $N=5$. For the TM setup we chose the number of clauses to be 1000, the threshold at 100, and the specificity to 15. The max number of literals per clause was fixed at 50. These choices were made given the performances from the classification numerical experiments. They generally yielded top results.

\begin{figure}[ht]
    \centering
    \subfloat[Example 1 Harmonic series]{{\includegraphics[width=7cm]{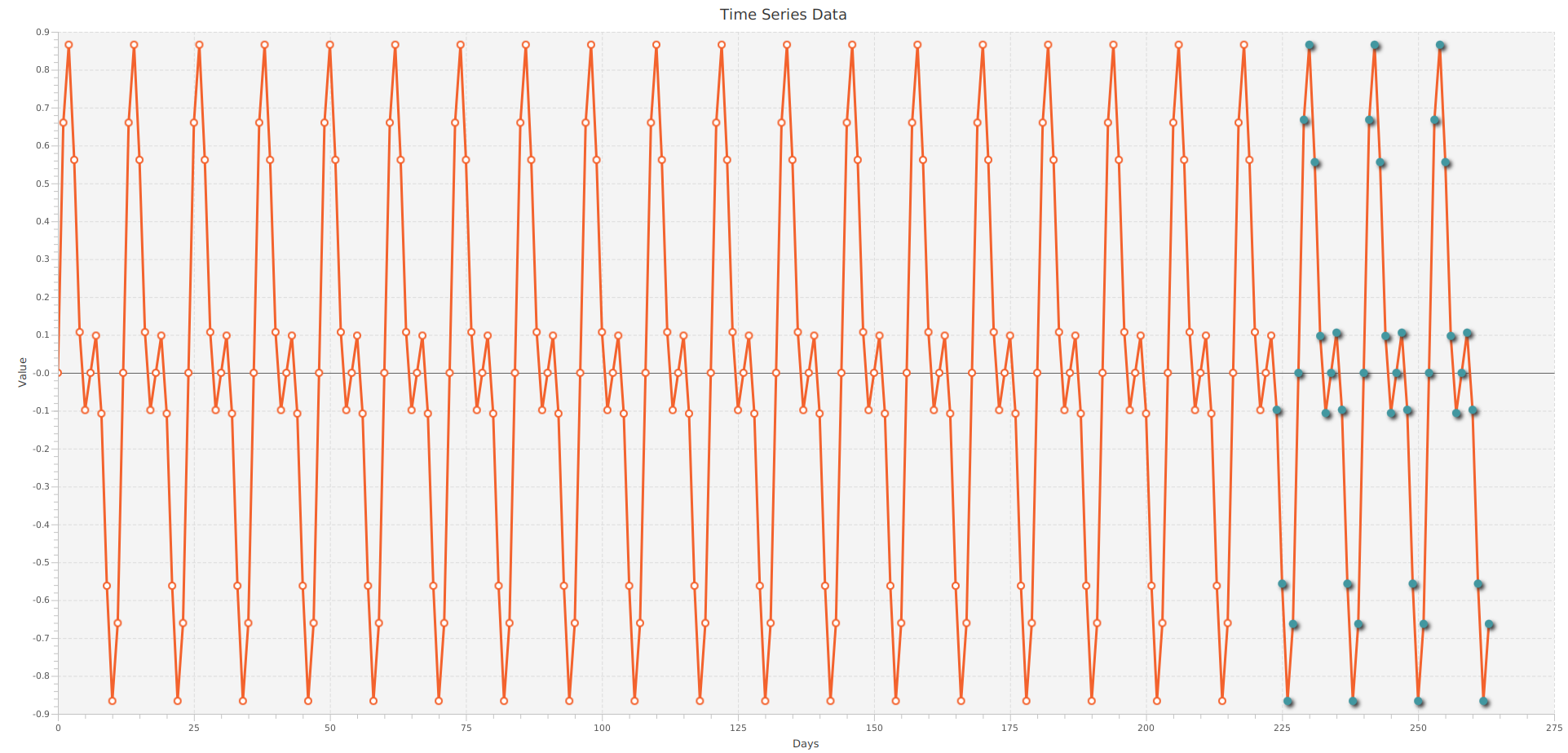} }}%
    \qquad
    \subfloat[Example 2 Harmonic series]{{\includegraphics[width=7cm]{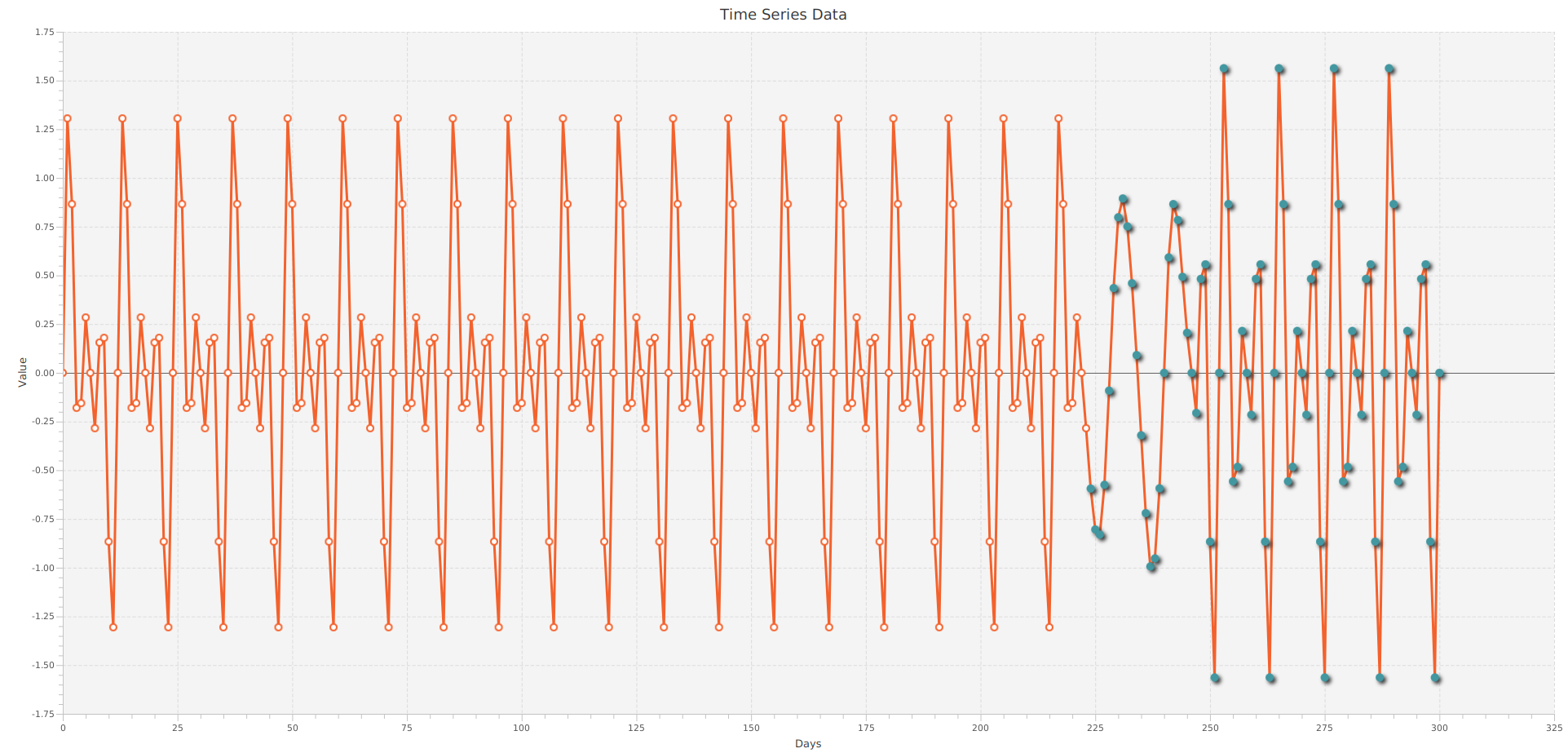} }}%
    \caption{Generating 24 steps ahead for two different deterministic series.}%
    \label{fig:example1harmonic}%
\end{figure}

\begin{figure}[ht]
    \centering
    \subfloat[Example 3 Harmonic series]{{\includegraphics[width=7cm]{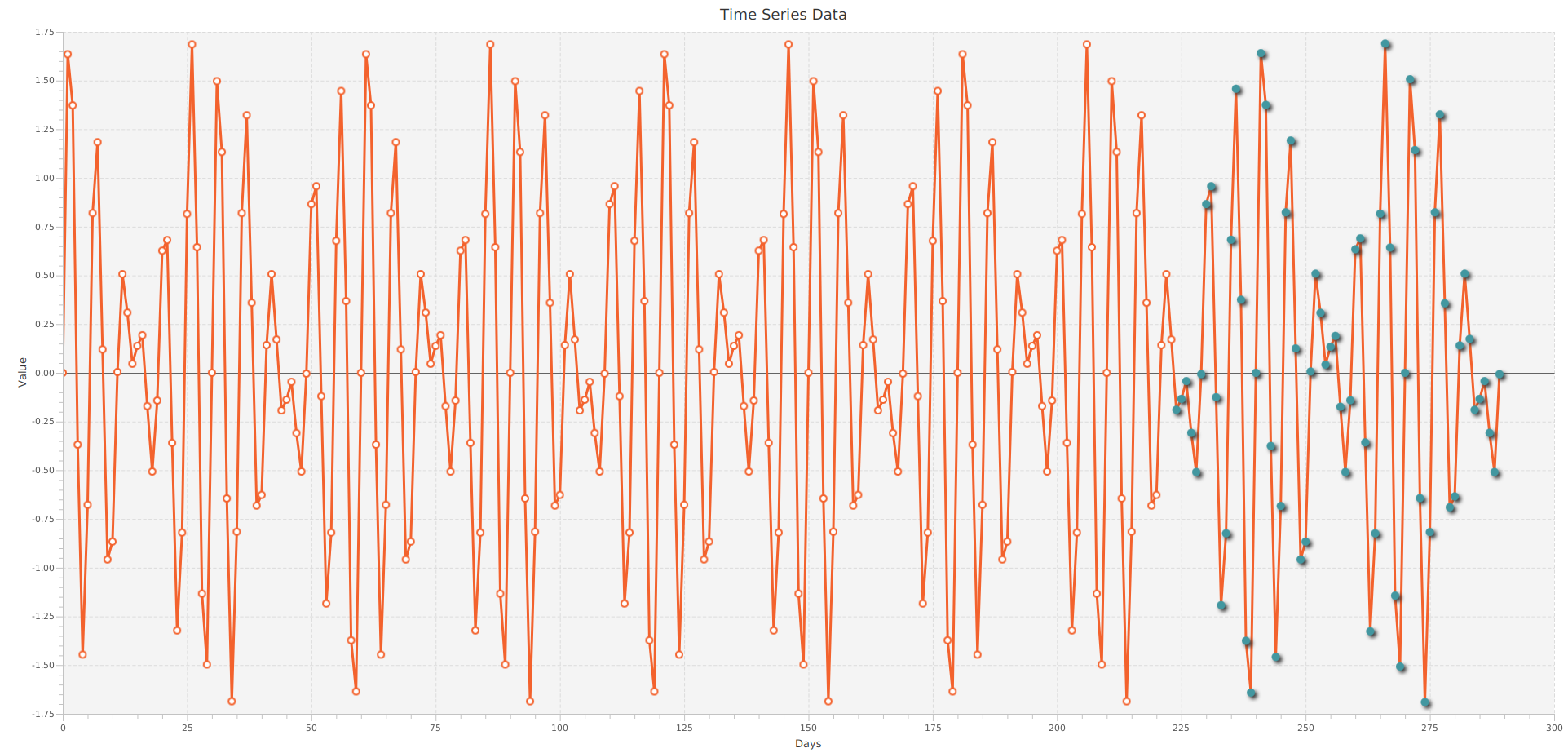} }}%
    \qquad
    \subfloat[Example 4 Harmonic series]{{\includegraphics[width=7cm]{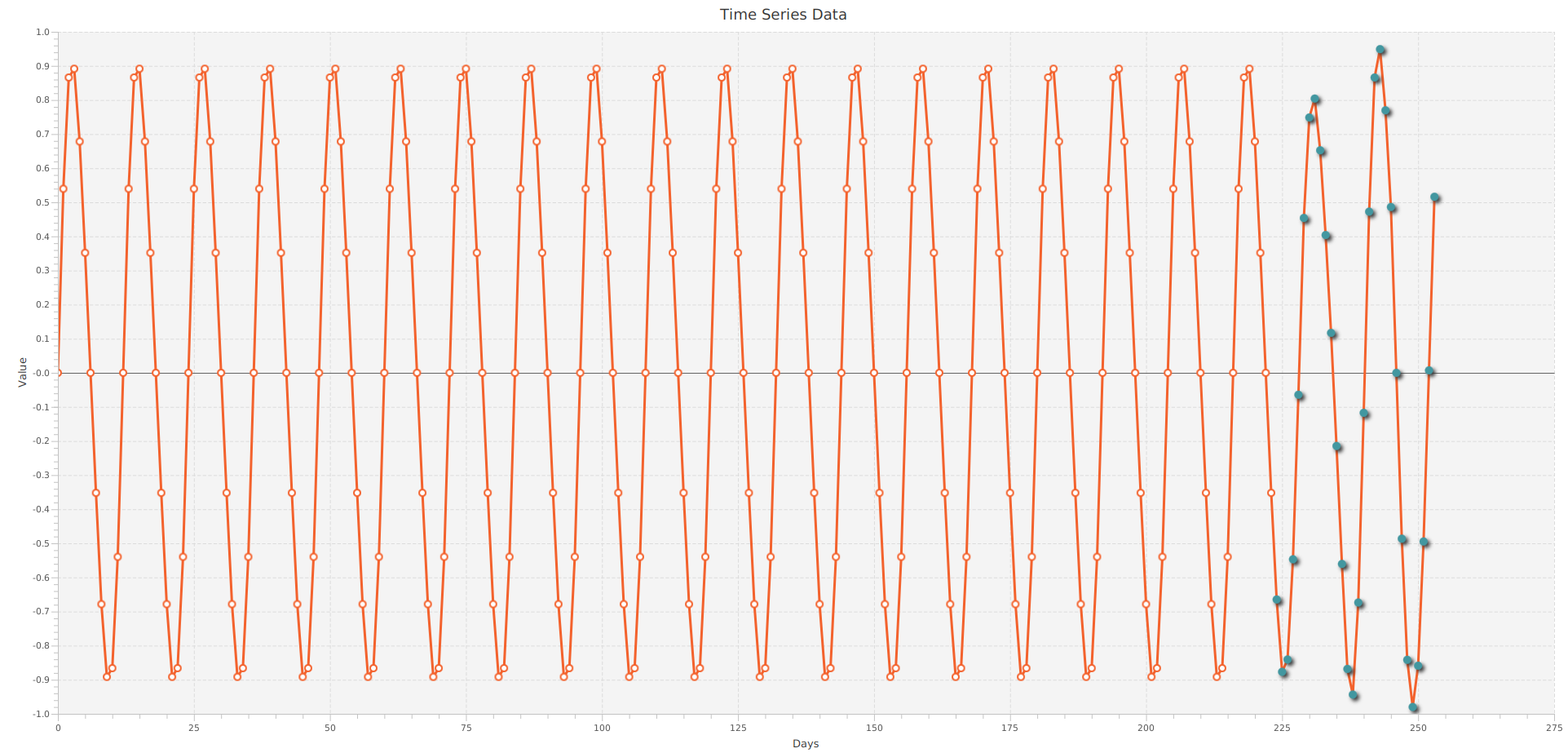} }}%
    \caption{Generating 24 steps ahead for two different deterministic series.}%
    \label{fig:example2harmonic}%
\end{figure}

Figures \ref{fig:example1harmonic} and \ref{fig:example2harmonic} show 4 examples of various harmonic series where we sample 220 observations. For each series the first 220 observations were encoded using N-Gram windowing technique shown in figure \ref{ref:forecasting} after which we then learned the samples from the encoded N-Grams and labeled them with the next values in the sequence, as shown in figure \ref{ref:hybridForecast}. 

We can see here that the hybrid forecasts are able to extract the harmonic structure well enough to generalize a reproduction of the periodicity involved. 

\begin{figure}[ht]
    \centering
    \subfloat[AR(1) series with AR coefficient .7]{{\includegraphics[width=7cm]{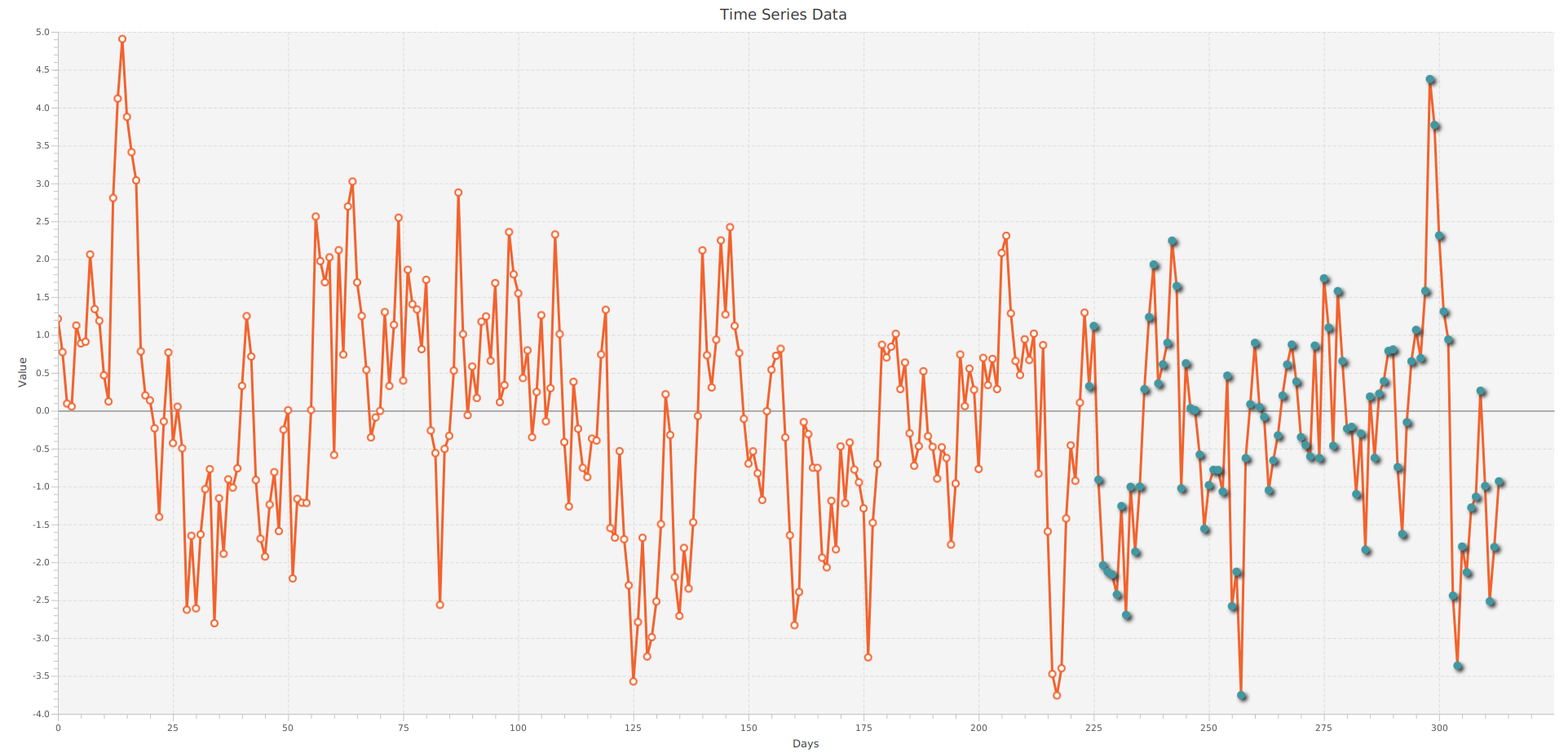} }}%
    \qquad
    \subfloat[AR(1) series with AR coefficient .4]{{\includegraphics[width=7cm]{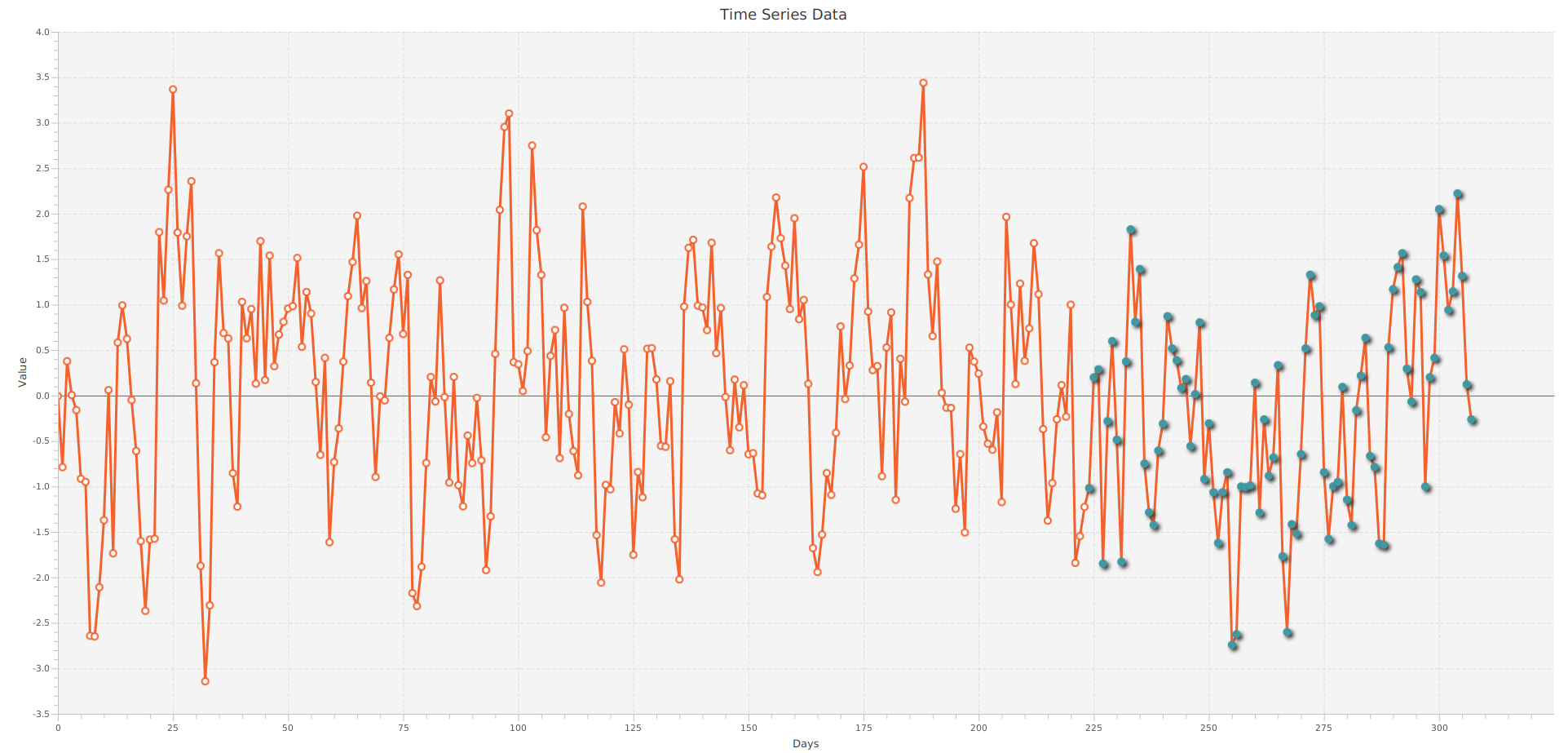} }}%
    \caption{Generating 24 steps or more ahead for two different stochastic series.}%
    \label{fig:example2ar}%
\end{figure}

Each new value generated is only dependent on the previous 4 values, since the N-Gram length is 5. Subsequent forecasts are then used to generate new values in the sequence. 

Next we expand on this by now simulating series from stochastic models. Here we consider a simple AR(1) model with very high AR coefficient, an ARMA(1,1) model. Finally in order to see how well it performs on seasonal data, we also sample data from a seasonal AR model with strong seasonal coefficients. 

\begin{figure}[ht]
    \centering
    \subfloat[ARMA(1,1) series .9 and -.1]{{\includegraphics[width=7cm]{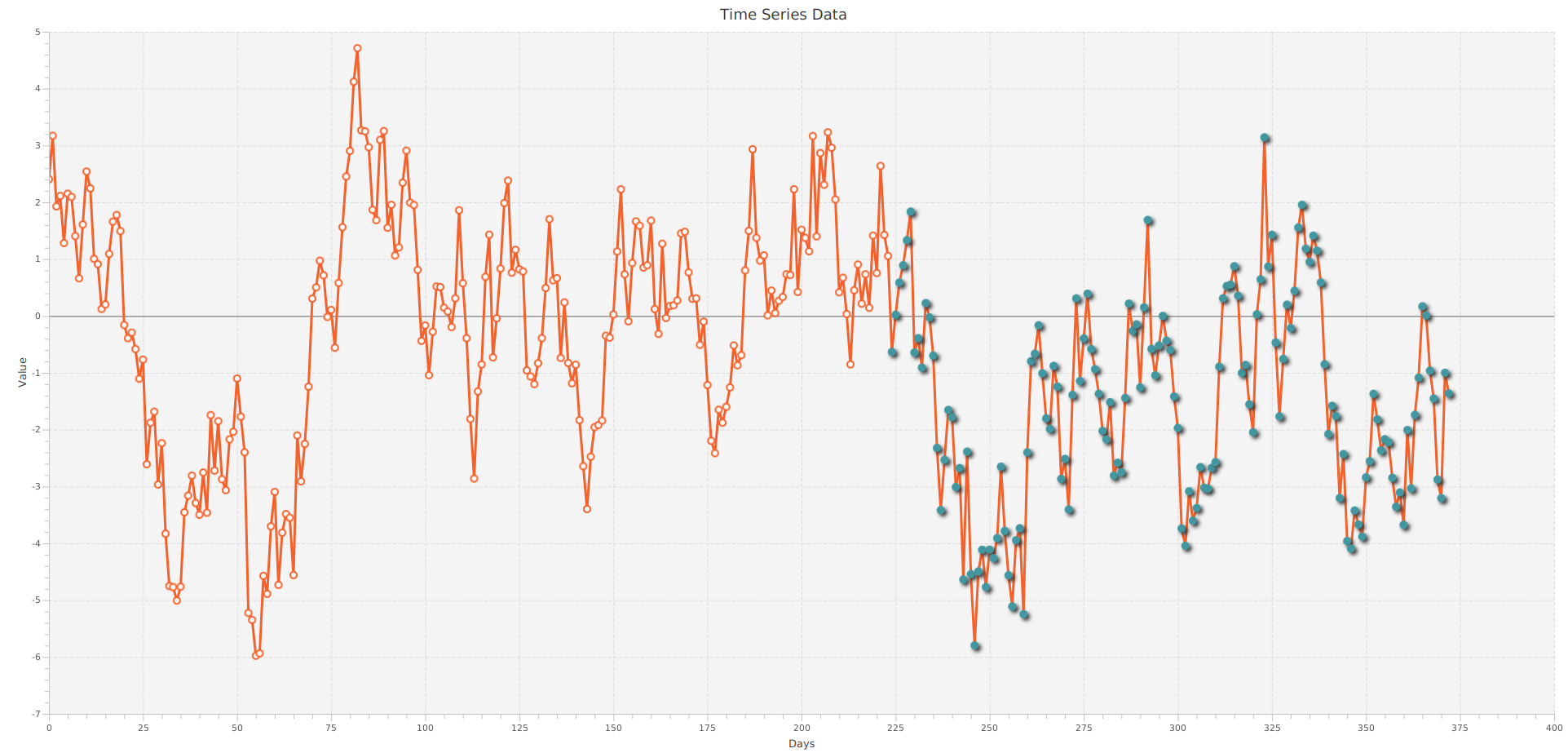} }}%
    \qquad
    \subfloat[ARMA(1,1) series .9 and -.1]{{\includegraphics[width=7cm]{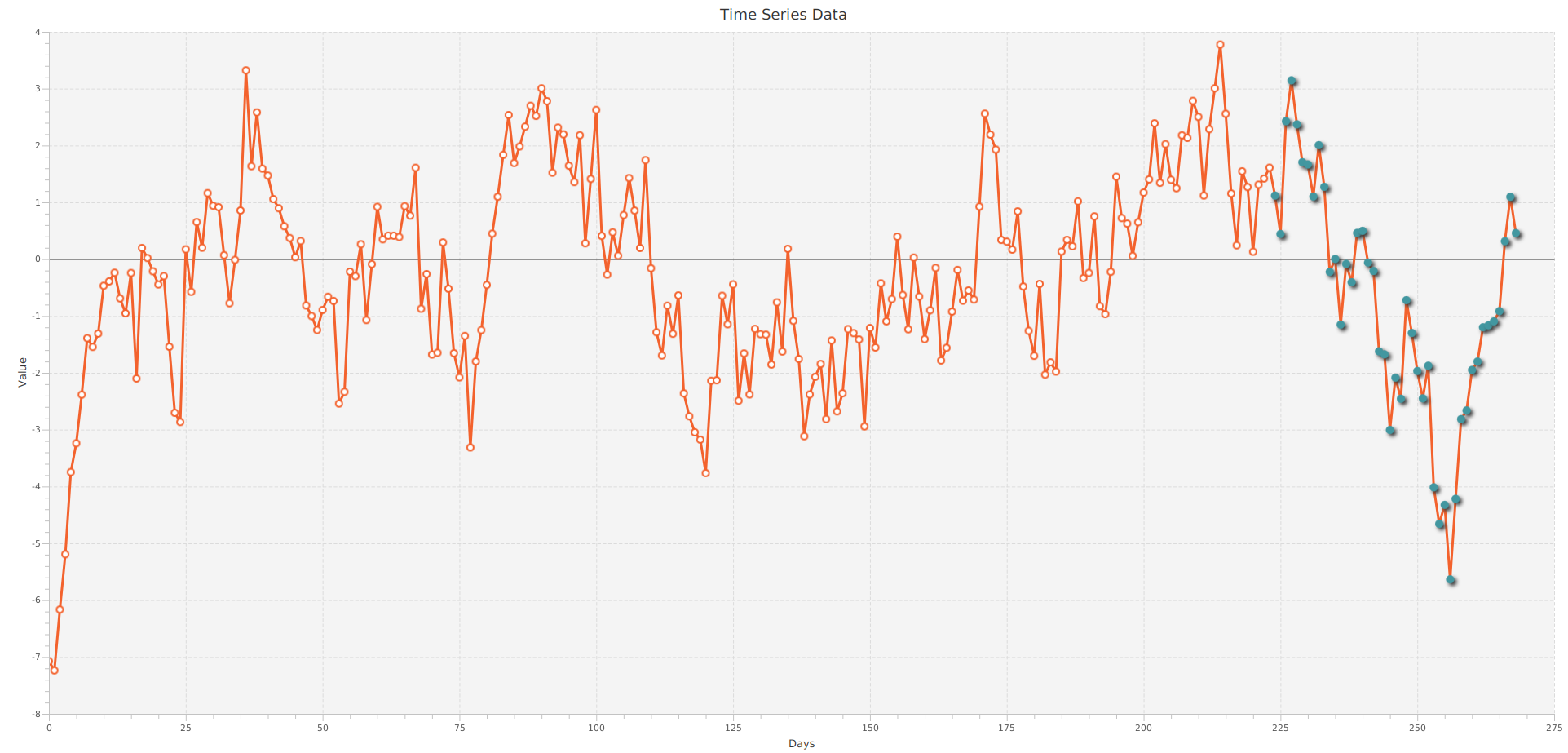} }}%
    \caption{Generating 24 steps or more ahead for two different stochastic series.}%
    \label{fig:example2arma}%
\end{figure}

\begin{figure}[ht]
    \centering
    \subfloat[SAR(1) series with seasonal AR coefficient .7]{{\includegraphics[width=7cm]{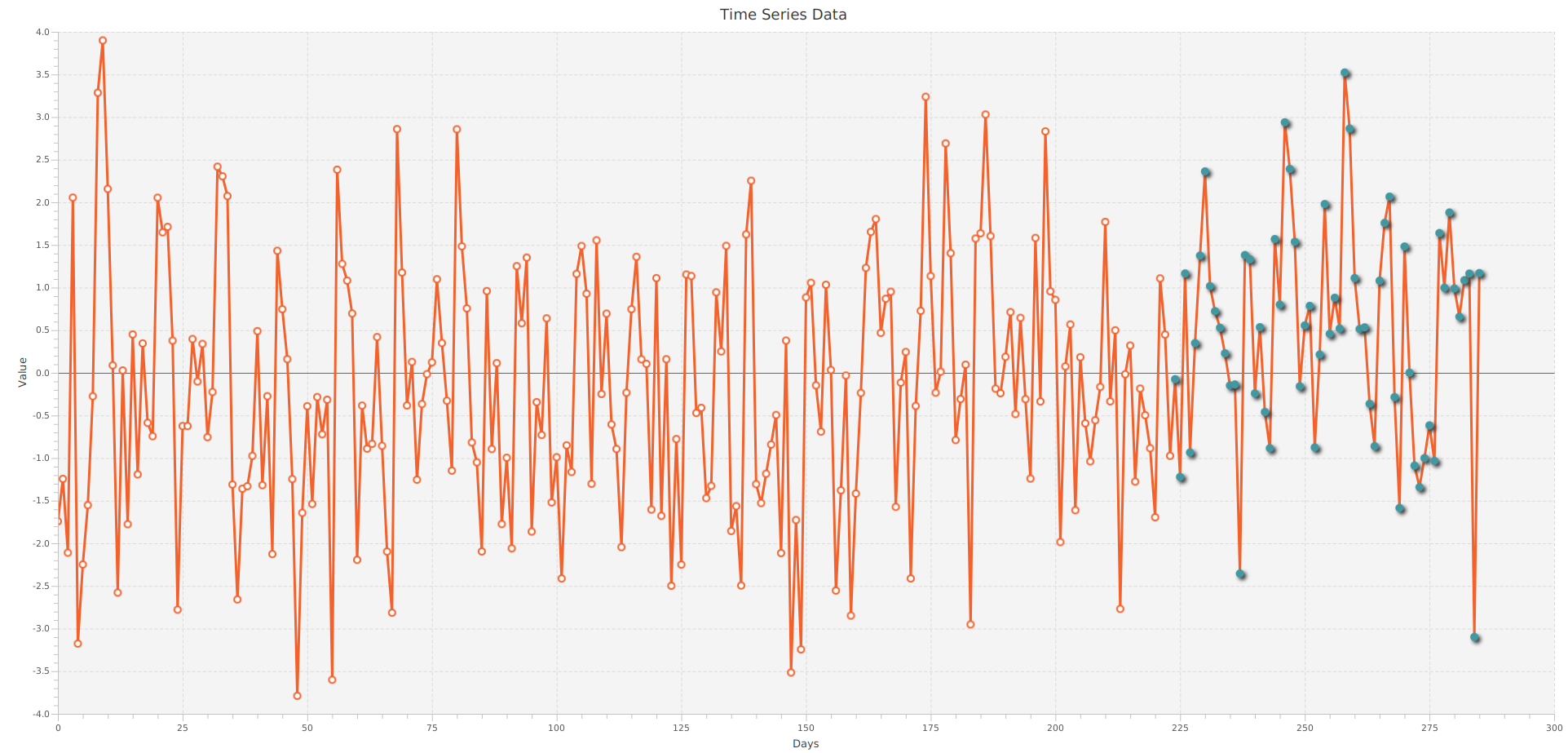} }}%
    \qquad
    \subfloat[SAR(1) series with seasonal AR coefficient .4]{{\includegraphics[width=7cm]{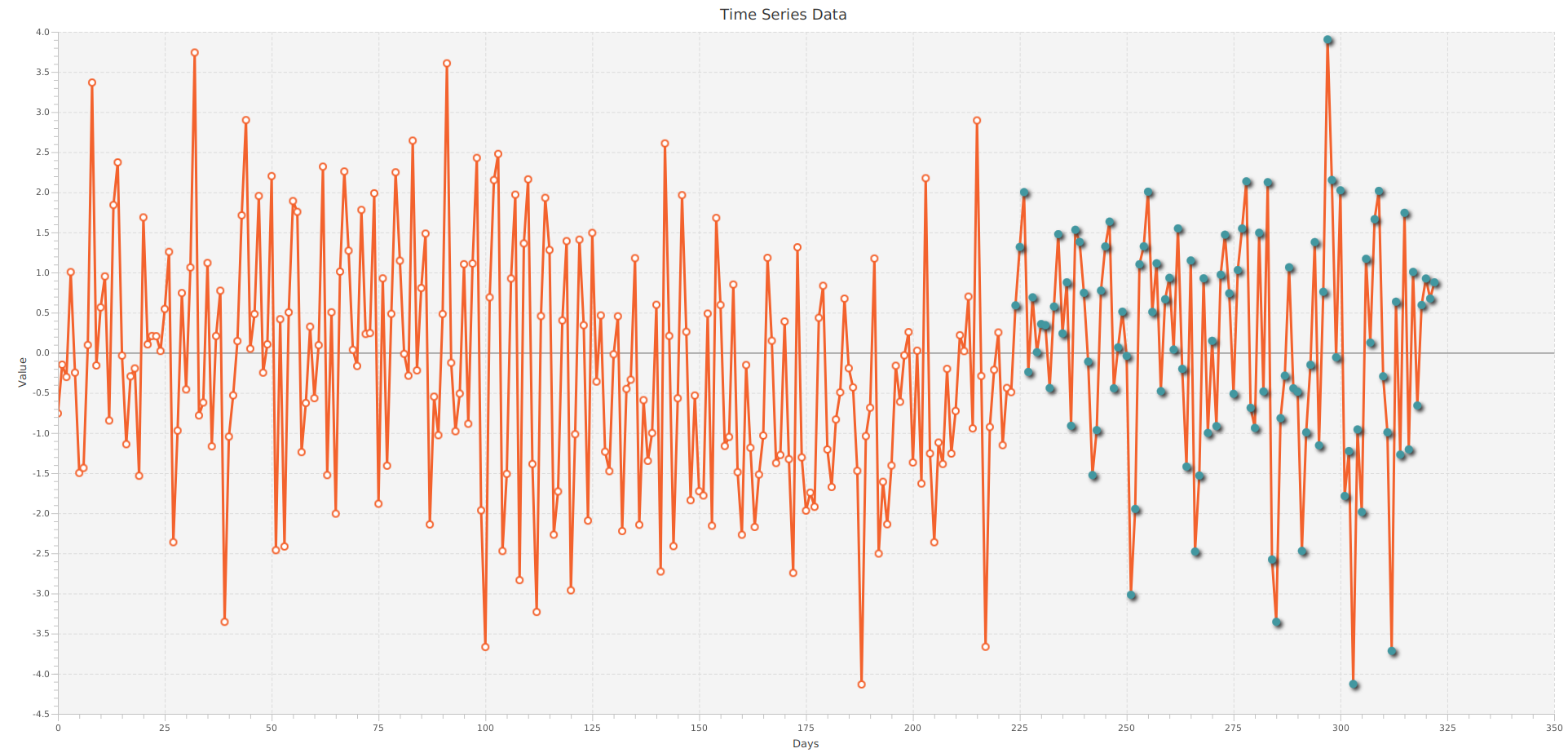} }}%
    \caption{Generating 24 steps or more ahead for two different stochastic series.}%
    \label{fig:example2sar}%
\end{figure}

A seasonal model produces new challenges not seen in AR(MA) models as the sequence generator also has to "learn" what frequency the seasonality occurs at. We can see in \ref{fig:example2sar} that is able to find estimate the seasonality fairly well. 

To generate empirical results on the forecasting ability of the approach, we sample from each model 50 times and generate sequentially a 24-step ahead forecast. For each forecast, we compute the error with the ground truth next value in that sequence. The time series have been normalized between [-1,1] to ensure normalized regular record keeping of the errors (so for example, in an extreme case, a mean error of $\sqrt(48)/24$ would imply all ground truth values were 1 but forecasts were -1, which gives an upper bound on the errors.)

\begin{table}[h!]
\centering
\begin{tabular}{l c c c c c}
\toprule
\textbf{Model} & \textbf{Coefficients} & \textbf{N-Gram 3} & \textbf{N-Gram 5} & \textbf{N-Gram 7} \\
\midrule
Harmonics     &            & 0.078 (0.015) & 0.044 (0.012) & 0.051 (0.022) \\
AR(1)         & 0.2        & 0.298 (0.020) & 0.102 (0.020) & 0.143 (0.052) \\
AR(1)         & 0.4        & 0.245 (0.114) & 0.152 (0.085) & 0.215 (0.131) \\
AR(1)         & 0.7        & 0.245 (0.011) & 0.140 (0.081) & 0.200 (0.121) \\
SAR(1)        & 0.1; 0.7   & 0.511 (0.315) & 0.312 (0.181) & 0.415 (0.317) \\
SAR(1)        & 0.3; 0.7   & 0.421 (0.273) & 0.552 (0.121) & 0.651 (0.323) \\
ARMA(1,1)     & 0.7; -0.1  & 0.245 (0.014) & 0.152 (0.074) & 0.201 (0.027) \\
\bottomrule
\end{tabular}
\caption{Table of Model Coefficients and N-Gram Values with corresponding standard errors}
\label{tab:model-table}
\end{table}

We can see that the forecasting performs best on the deterministic Harmonic series, which is to be expected. the standard error is near 7 percent over a 24 step forecast, for 3 N-Gram encoding, and around 4 percent for 5 N-Gram encoding. In general, the best performance comes from 5 N-Gram encoding which is able to capture best the periodicity.  For the AR models, we see they are fairly consistent regardless of the coefficient, with a rather tight standard deviation. 

The seasonal AR model poses more challenges because not only does it have to correctly forecast the next value, but also determine what what frequency the seasonality is occurring. Naturally, with the larger variance at seasonal periods, the standard error will typically be larger. For both seasonal models, it still performs quite well with a "best" performance at around 31 percent error. Lastly, for the ARMA model the forecasting performance seems to be nearly as good as the standard AR model with a lower standard error near 15 percent.

\section{Conclusion}

In this paper we proposed a straightforward approach to combining the strengths of HVC with the machine learning power of Tsetlin machines to give a flexible approach for learning sequences, classifying them, and generating new ones. Our approach relied heavily on encoding procedures for sequences where we took advantage of a powerful N-Gram structure for spatialtemporal learning. Since the resulting HVs can be decoded as well, combined with the innate interpretable structure of Tsetlin machines, this hybrid approach could be an attractive alternative to larger models such as Deep learning sequence models (LSTMs, RNNs) which rely heavily on large scale weight estimation via matrix vector multiplication, backpropagation, optimization and parameter hypertuning. 

\subsection{Model Size}

This leads us to an additional note that we have not discussed in the paper but would be valuable for future work. Model size and computation learning in terms of energy expenditure could be optimized and improved greatly by taking advantage of the fact we are using binary vectors. The size of a model based on HVs depends on several factors, including the dimensionality of the vectors, the number of vectors used in the model (number of N-Grams and interval embedding vectors), but we can safely conclude that reasonably, with a dimension of 10k for or HV system, and with a thousand vectors, we have a size of 1,220KBs total (10,000bits/8=1,250bytes times 1000). Expanding to even larger dimensions, say D = 100k, we still only need a few MBs of chip memory. Now coupling with the TM, say 1000 clauses, each with max 32 bits for each automata, we are looking at a trivial amount of additional memory. 

\subsection{Future work}

Many direction exist in terms of where these models could be challenged next. The first area of improvement would be in understanding better the underperformance of some of the classification results, including when many classes are present. Here, it is of the author's hypothesis that simply a higher dimension, along with a more careful pruning and selection of model criterion could be achieved. Perhaps even varying the number of N-Grams, to see if they play a large role for a larger number of classes. Since no optimization was done in choosing the TM parameters, this would also be an area to look at more. Since the goal was to see if the approach could work "straight-out-of-the-box", we wanted to have the least amount of tinkering as possible in model architecture setups. 

The second area of work would be into taking a deeper dive into forecasting real-world series with more challenging dynamics. For example, some macroeconomic data, central orderbook data, and asset returns from the financial world, or signal sequences from the medical field. 

A third area of work would be to take into account additional time series to model and forecast mulitvariate time series. Here, the attraction would be to incorporte more features from either the underlying series (for example, volatility, seasonality), and see how classification and forecasting can be tackled when additional features are included.

\bibliographystyle{unsrt}  
\bibliography{references}

\end{document}